\documentclass[letterpaper]{article} 
\usepackage{aaai17}  
\usepackage{times}  
\usepackage{helvet}  
\usepackage{courier}  
\usepackage{url}  
\usepackage{graphicx}  
\usepackage[usenames,dvipsnames]{xcolor}
\begin{document}
\newcommand{\ce}[1]{\texttt{\small{#1}}}

\title{An Experimentation Platform for Explainable Coalition Situational Understanding}
\author{Katie Barrett-Powell, Jack Furby, Liam Hiley, Marc Roig Vilamala \\ {\bf \Large Harrison Taylor, Federico Cerutti, Alun Preece} \\
Cardiff University\\
Cardiff, UK
\AND 
Tianwei Xing, Luis Garcia, Mani Srivastava \\
University of California \\
Los Angeles, USA
\And 
Dave Braines \\
IBM Research Europe \\ 
Hursley Park, Hampshire, UK}

\newcommand{\aedit}[1]{{\color{red} #1}}

\maketitle

\vspace{1cm}

\begin{abstract}
We present an experimentation platform for coalition situational understanding research that highlights capabilities in explainable artificial intelligence/machine learning (AI/ML) and integration of symbolic and subsymbolic AI/ML approaches for event processing. The Situational Understanding Explorer (SUE) platform is designed to be  lightweight, to easily facilitate experiments and demonstrations, and open. We discuss our requirements to support coalition multi-domain operations with emphasis on asset interoperability and ad hoc human-machine teaming in a dense urban terrain setting. We describe the interface functionality and give examples of SUE applied to coalition situational understanding tasks.
\end{abstract}


\section{Introduction}

A key characteristic of multi-domain operations (MDO)~\cite{TRADOC:2018} is that near-peer adversaries will be contesting all domains --- from dense urban terrain to space and cyberspace --- wherein those adversaries attempt to achieve \emph{stand-off} by reducing allies' speed of recognition, decision and action, as well as by fracturing alliances through multiple means: diplomatic, economic, conventional and unconventional warfare, including information warfare. A critical requirement for allies is rapid and continuous integration of capabilities to collect, process, disseminate and exploit actionable information and intelligence. To achieve this, the MDO \textit{layered ISR} concept envisions exploitation of `an existing intelligence, surveillance, and reconnaissance (ISR) network developed with partners\ldots that consists of overlapping systems of remote and autonomous sensors, human intelligence, and friendly special operations forces' (\cite{TRADOC:2018}, pp.33--34). Maximally utilising ISR assets in the contested MDO environment requires an ability to share resources among coalition partners in an open but controlled environment, with knowable levels of trust and assurance. Our research is focused on three facets of MDO~\cite{Preece:2019}:

\textbf{Enhanced asset interoperability} Here we focus specifically on assets based on artificial intelligence (AI) and machine learning (ML) technologies, both subsymbolic (e.g., deep neural networks) and symbolic (e.g., logic-based approaches).  To achieve interoperability between these assets --- as well as human assets --- in a coalition MDO setting, we need subsymbolic AI/ML agents to be able to share uncertainty-aware representations of insights and knowledge that can then be communicated to humans and symbolic AI/ML assets; and we need to enable symbolic AI assets --- and humans --- to share inferences and insights to subsymbolic AI/ML agents.

\textbf{Human-machine teaming} Specifically we focus on \emph{ad hoc coalition teams} involving humans and hybrid (subsymbolic and symbolic) AI/ML assets where there is a key need to rapidly build trust and confidence. Therefore, our work seeks to advance capabilities in explainable AI/ML to allow a human operative to `calibrate their trust' in an AI/ML asset potentially provided by a different coalition partner~\cite{Tomsett:2020}. The purpose of human-machine teaming is to aim for each party to exploit the strengths of, and compensate for the weaknesses of, the other~\cite{Cummings:2014}. 

\textbf{Dense urban terrain analytics} Accelerating global rates of urbanisation and the emerging strategic importance of cities/megacities means that MDO operations will take place within dense urban terrain, giving rise to specific physical, cognitive, and operational characteristics. ISR will exploit and augment civilian infrastructure, for example, civilian closed circuit television (CCTV) cameras, augmented with processing for the detection and tracking of activities, and to support building pattern-of-life models. ISR tasks cannot necessarily plan in advance what collection and processing will be needed: analytics composition will be dynamic and context-specific, with continual re-provisioning and resource optimisation~\cite{White:2019}.

\begin{figure}[t]
\centering
\includegraphics[width=0.35\textwidth]{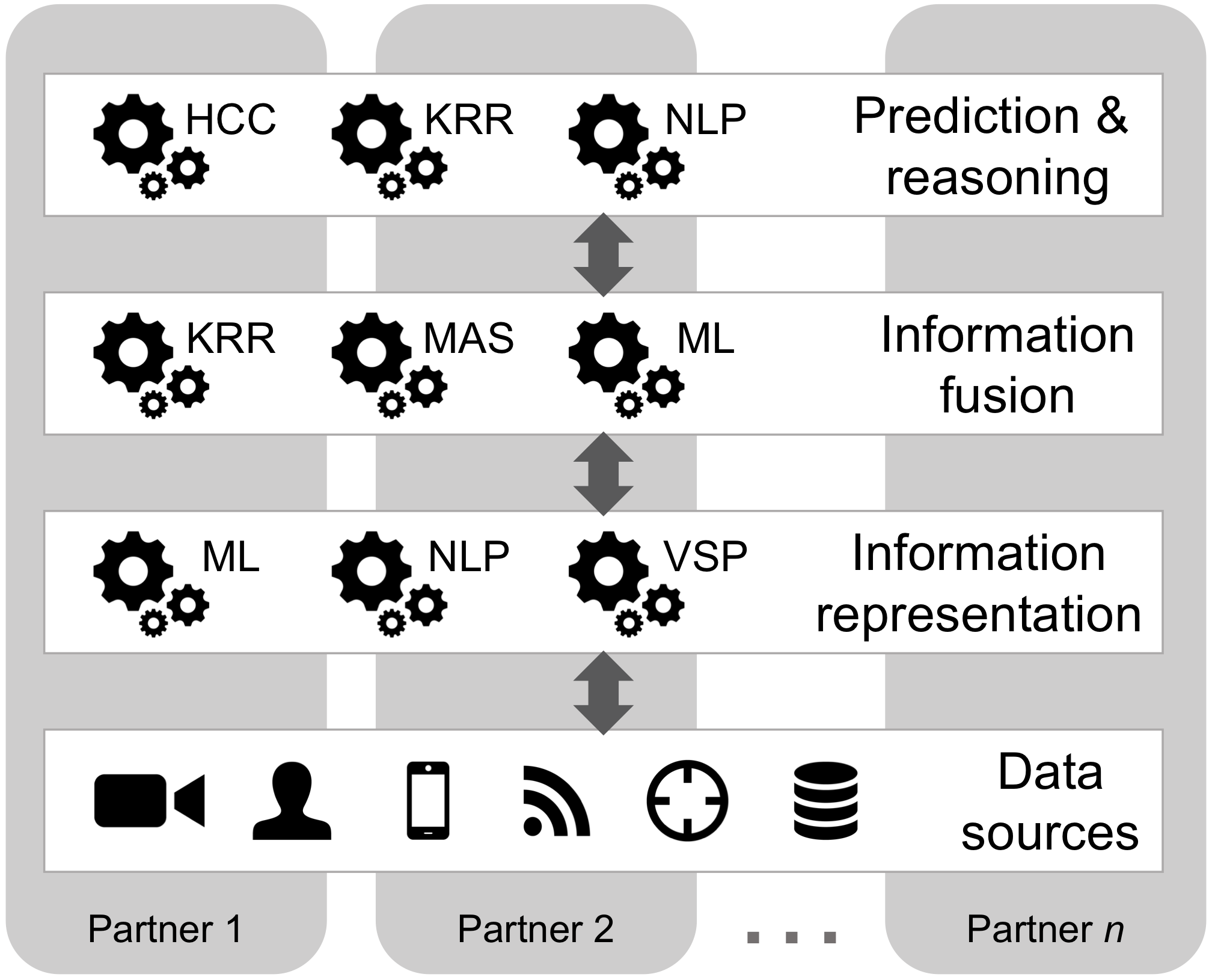}
\caption{CSU layered model (from~\cite{Preece:2017}) distributed across multiple partners and employing multiple technologies: human-computer collaboration (HCC), knowledge representation \& reasoning (KRR); multi-agent systems (MAS); machine learning (ML); natural language processing (NLP), vision and signal processing (VSP).}
\label{fig:layers}
\end{figure}

Our concept of \emph{coalition situational understanding} (CSU)~\cite{Preece:2017} extends the notion of situational understanding (the `product of applying analysis and judgment to the unit's situation awareness to determine the relationships of the factors present and form logical conclusions concerning threats to the force or mission accomplishment, opportunities for mission accomplishment, and gaps in information'~\cite{Dostal:2007}) to a coalition MDO context, specifically featuring by a layered architecture consisting of sensing, information representation and fusion, and prediction/reasoning assets (Figure~\ref{fig:layers}). Assets at all layers are distributed across the network and may be provided by any coalition partner. Moreover, CSU emphasises real-time analytics and temporal information processing, exploitation of multimodal data sources (e.g., imagery, acoustic, open source media), and the decentralised placement of assets at or near the network edge.

The focus of this paper is on an experimentation platform for our CSU research, that specifically highlights capabilities in (1) explainable AI/ML for asset trust calibration and (2) integration of symbolic and subsymbolic AI/ML assets for event processing. This \emph{Situational Understanding Explorer} (SUE) platform has been designed to be very lightweight (to more easily facilitate experiments and demonstrations) and open\footnote{Open source: https://github.com/KBarrett-Powell/SUE}. The next section discusses the requirements for SUE; subsequent sections describe the interface functionality and give examples of SUE applied to CSU tasks in line with (1) and (2) above. 

\section{SUE Requirements}

This section summarises the main requirements for SUE, which map to the main functional elements of the platform. 

\textbf{Sensors} Corresponding to the data source layer in Figure~\ref{fig:layers}, SUE must represent sensors of various kinds placed in the physical environment. Sensors need to be associated with (owned by) multiple coalition partners. As noted above, our focus is on dense urban settings. Moreover, our initial focus has been on 2D representations given the ease of acquiring open source 2D mapping data; a 3D environment is under exploration. 

\textbf{Events} As the main product of the AI/ML services in the higher layers in Figure~\ref{fig:layers}, we focus on the detection of, and reasoning about, events. Events involve objects and actors, and are situated in time and space. A \emph{simple event} is an event detected by a single sensor; a \emph{complex event} is a spatiotemporal combination of multiple simple events from one or more sensors. SUE must visualise events in time and space and provide details of their components (e.g., objects, actors) and relationships (e.g., between simple and complex events).

\textbf{Explanations} The AI/ML services need to be capable of providing explanations for their detection and reasoning decisions with respect to events. In previous work we assert that explanations should be layered to provide distinct kinds of information relating to system verification (that the decision was reached in the correct way) and validation (that it's the correct decision)~\cite{Preece:2019}. SUE needs to be open to plugging-in these different elements of explanations.

\textbf{Uncertainty management} A key element of our work in trust calibration is the management of uncertainty, including distinctions between aleatoric and epistemic uncertainty~\cite{Tomsett:2020}. Again, SUE needs to be open to visualising uncertainties generated by the various kinds of AI/ML model in relation to detection and inference of events.

\textbf{Human-machine collaboration} In relation to our need to explore human-machine teaming, especially where human operatives may have limited technical training, we need interface affordances that are widely familiar, e.g., via smartphone apps. For the initial version of SUE we focused on direct manipulation of map-based and visual elements, with control also afforded by a conversational interface.

\textbf{Online and offline modes} To support experimentation and demonstration, SUE is required to work both offline, displaying a pre-prepared temporal stream of events, or online, receiving `live' events.

\vspace{0.25cm}

\begin{figure*}[tb]
\centering
\includegraphics[width=0.8\textwidth]{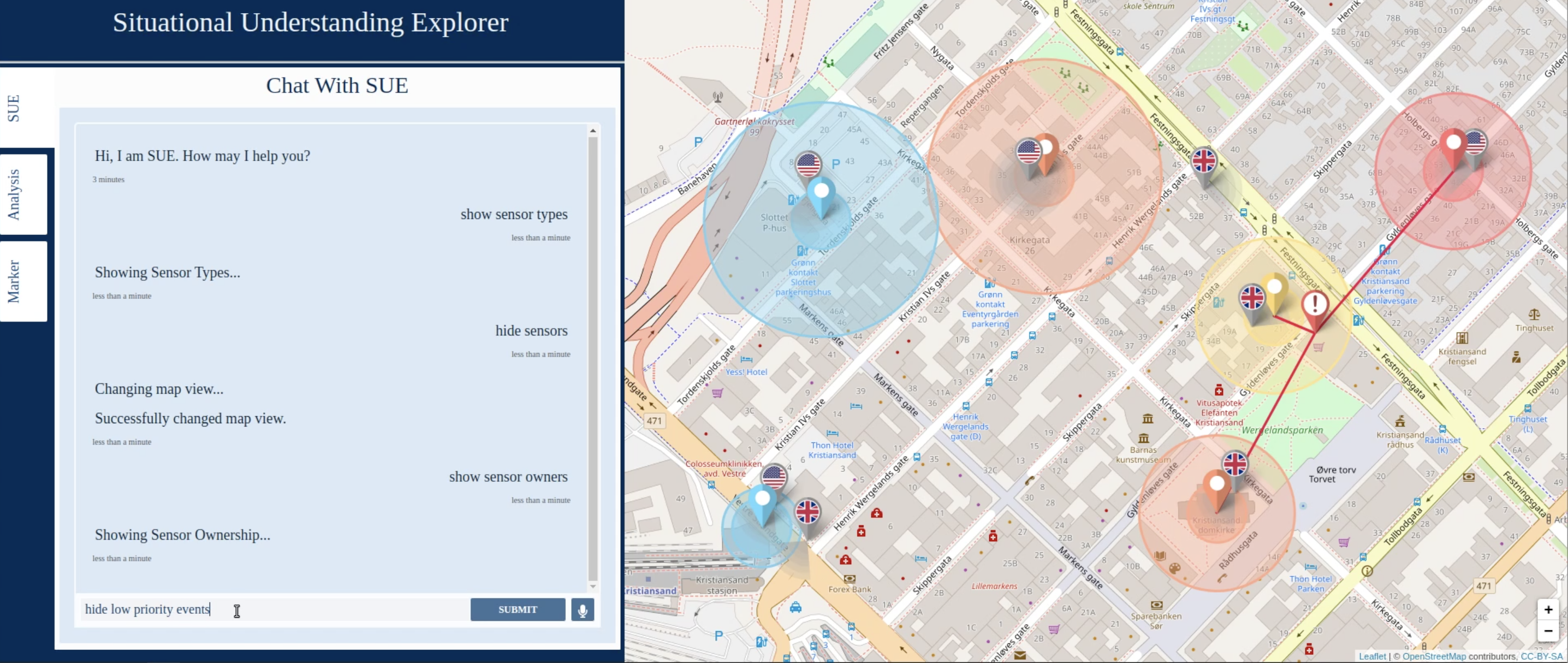}
\caption{The SUE interface showing the conversational panel (left) and the sensor m ap view (right) with sensor ownership shown via the `flag' icons.}
\label{fig:sue1}
\end{figure*}

\noindent
In developing the platform to meet these requirements, we considered building upon a number of pre-existing works. The NodeRED dashboard\footnote{https://flows.nodered.org/node/node-red-dashboard} offers convenient Internet of Things (IoT) sensor integration and simple events. We opted not to build on this since its emphasis is on lower-level service interoperability rather than higher-level situational understanding functionality; however, we designed SUE to interoperate with NodeRED via its online mode. We also considered using an open knowledge graph-based platform, Cogni-Sketch~\cite{Braines:2020}, which is well-suited to the higher-level situational understanding functions but not geared towards handling event streams. Again, we integrated SUE with knowledge graph functionality via linking to Cogni-Sketch.

\section{Overview of SUE and Examples}

This section presents a brief overview of the platform, its interface and features. SUE is implemented in JavaScript on Node.js\footnote{https://nodejs.org/}, with LeafletJS\footnote{https://leafletjs.com} providing map integration using OpenStreetMap\footnote{https://www.openstreetmap.org} map tiles. The conversational user interface (CUI) functionality is provided via Rasa\footnote{https://rasa.com}.

\begin{figure}[tb]
\centering
\includegraphics[width=0.4\textwidth]{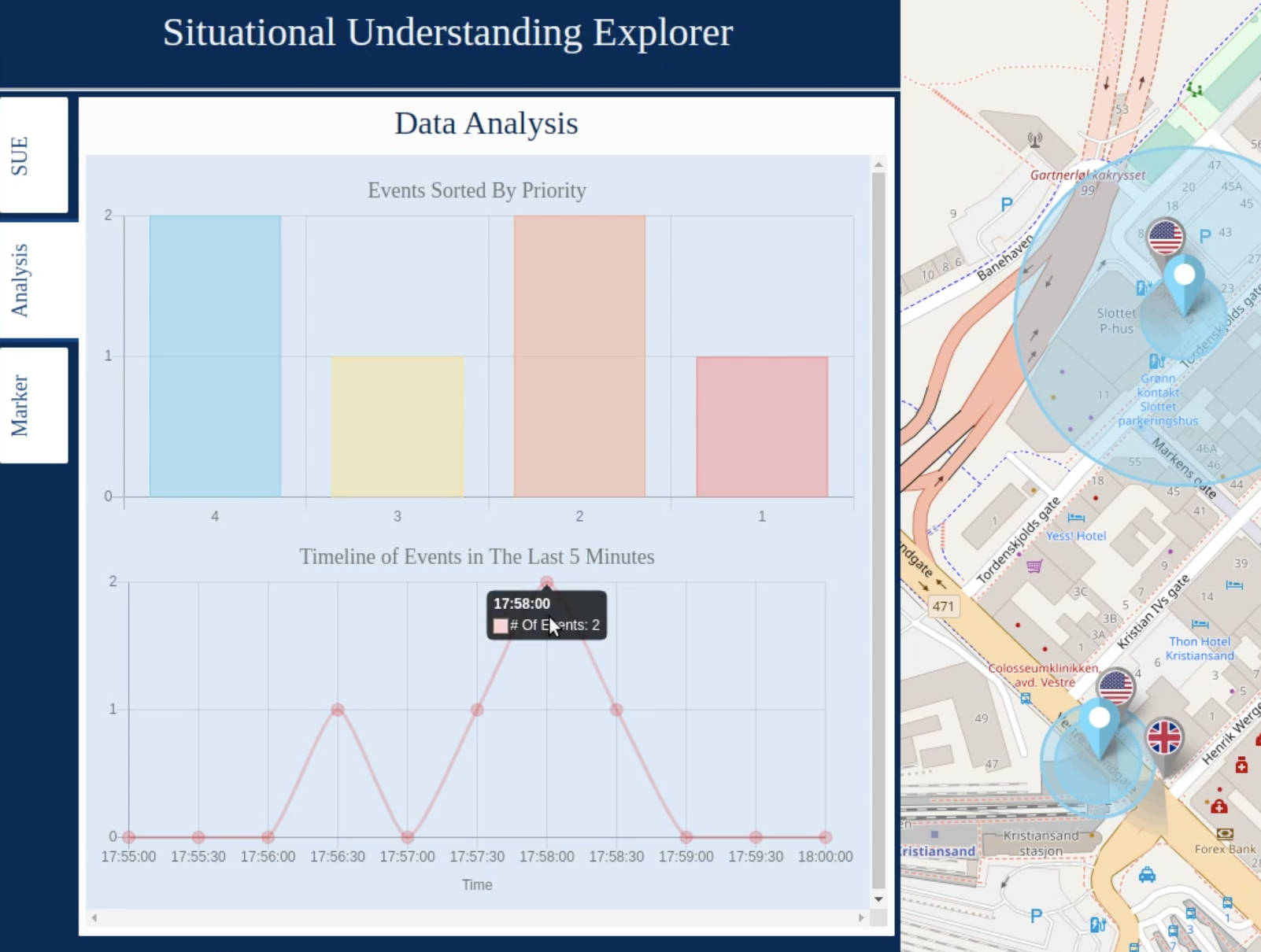}
\caption{The SUE interface analytics panel (left) showing summaries of events and event types.}
\label{fig:sue2}
\end{figure}

Figure~\ref{fig:sue1} shows the SUE interface with tabs on the left for the CUI, analytics panel and event marker panel. The map panel shows sensor and event positions. Each sensor can be shown either as its type (camera, microphone, etc) or coalition owner (seen here: USA and UK). Events are shown within coloured nested circles; in each case the outer circle approximates the region in which the event is located based on the sensed data, and the inner circle is always a constant size and serves to highlight the colour associated with the event, where the colours denote the degree of belief that something significant to the user is happening: red = strong, amber = medium, yellow = weak, blue = not significant. Sensor and event map markers are directly manipulable or controllable via the CUI: in this example we see the CUI being used to toggle between sensor type and owner views. Another CUI option allows the colours to be changed to more accessible variants for users with colour-blindness. Where simple events are related as part of a complex event (see the second example below for further details) they are joined by red lines to a complex event marker. Further details of events by type and by timeline are available on the analytics tab as shown in Figure~\ref{fig:sue2}.

The following subsections give two examples of how SUE emphasises our CSU research.

\paragraph{Multimodal Explanations}

\begin{figure}[tb]
\centering
\includegraphics[width=0.4\textwidth]{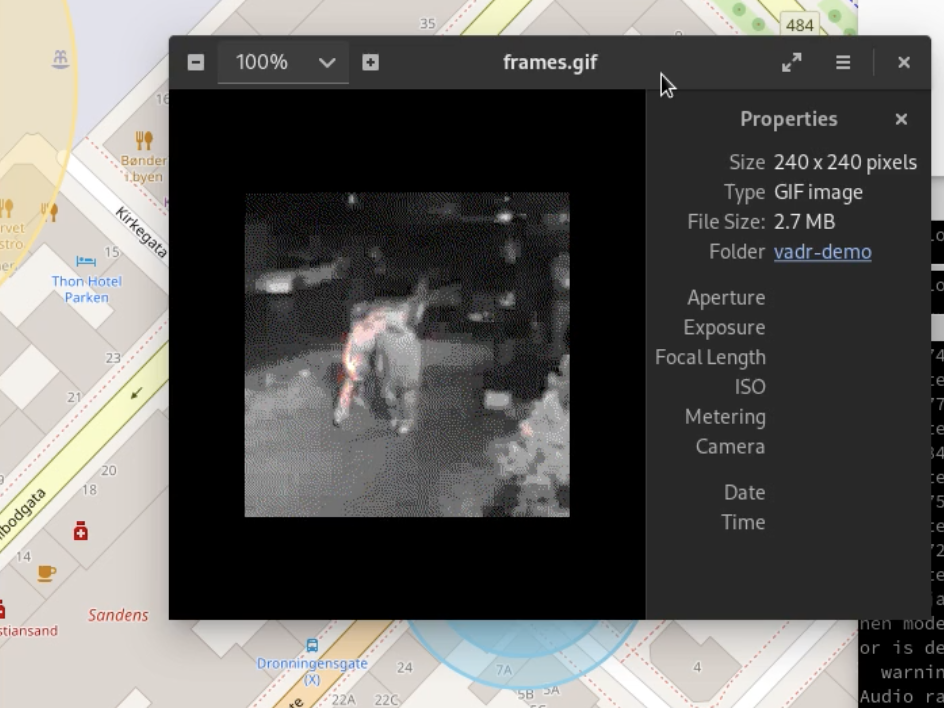}
\caption{A visual temporal relevance explanation shown in the SUE interface.}
\label{fig:savr1}
\end{figure}

Figure~\ref{fig:savr1} shows the detail associated with a simple event where a shooting has been detected in the city. This uses the selective relevance technique~\cite{Taylor:2020} to highlight features in the input that were most salient to the detection of the event. Here we can see red highlighting on the person to the left, who is the active shooter (the other person is merely walking past). The input here consists of both video and audio modalities and the selective relevance method allows us to determine which of the modalities is most salient to the event detection and, moreover, which are the key moments in the video and audio streams. The explanation shown in Figure~\ref{fig:savr1} has selected the most salient explanation as being visual.\footnote{Original video: https://www.youtube.com/watch?v=evgpWTm\_-rE --- the gunshots are heard out-of-context which is why the audio is less salient to the detection decision.} 

\begin{figure}[tb]
\centering
\includegraphics[width=0.4\textwidth]{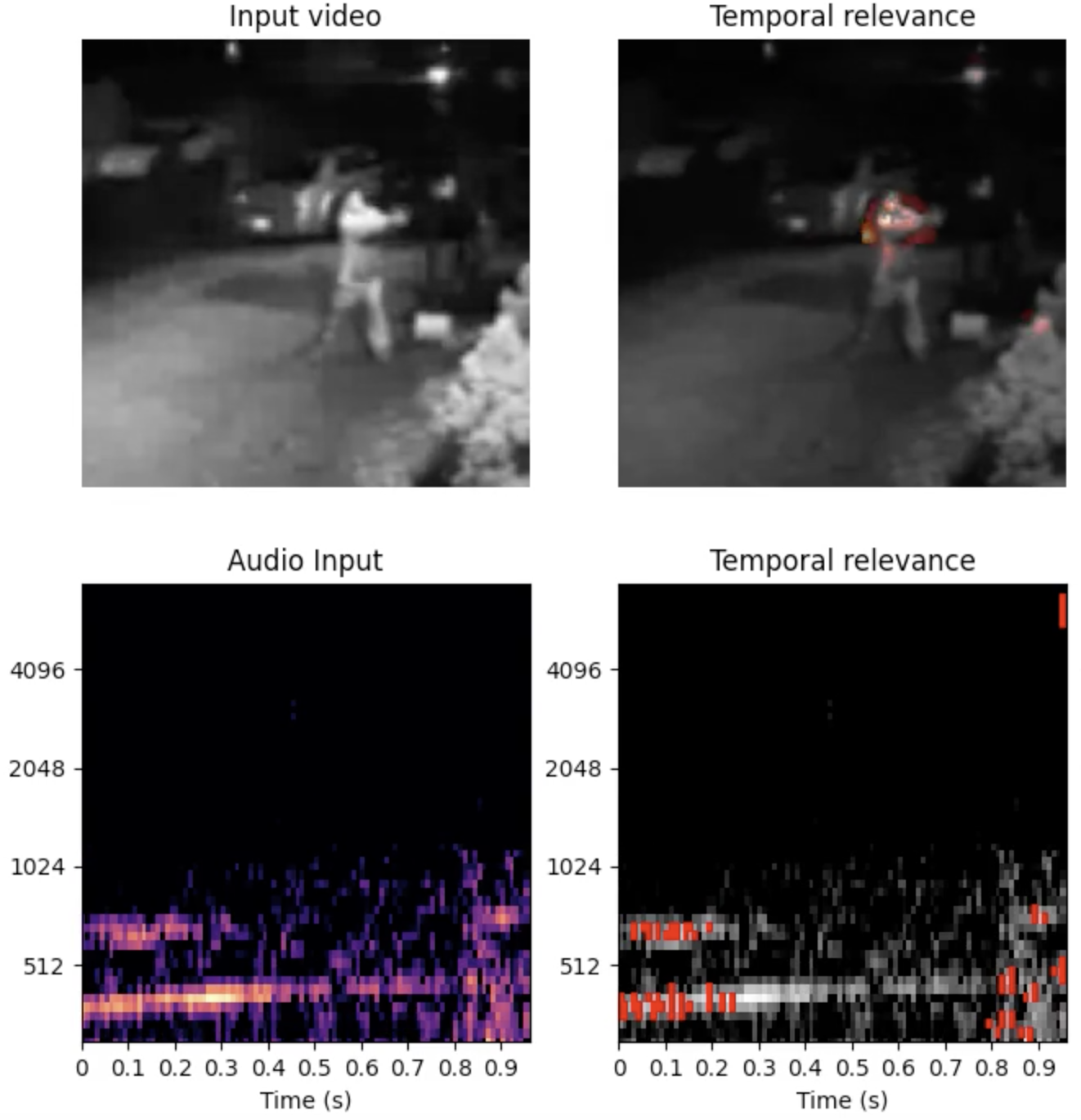}
\caption{Details of a multimodal selective relevance explanation showing visual and audio temporal relevance.}
\label{fig:savr2}
\end{figure}

The selective relevance method also separates temporal from spatial information in generating explanations. Figure~\ref{fig:savr2} shows an expanded view of this explanation where the original input --- video and audio in the form of a spectogram --- are on the left and the red-highlighted saliency regions are on the right. These frames are from a little later in the video where again we see highlights on the shooter, indicating temporal relevance (due to the person's movements).

\paragraph{Complex Events}

\begin{figure}[tb]
\centering
\includegraphics[width=0.4\textwidth]{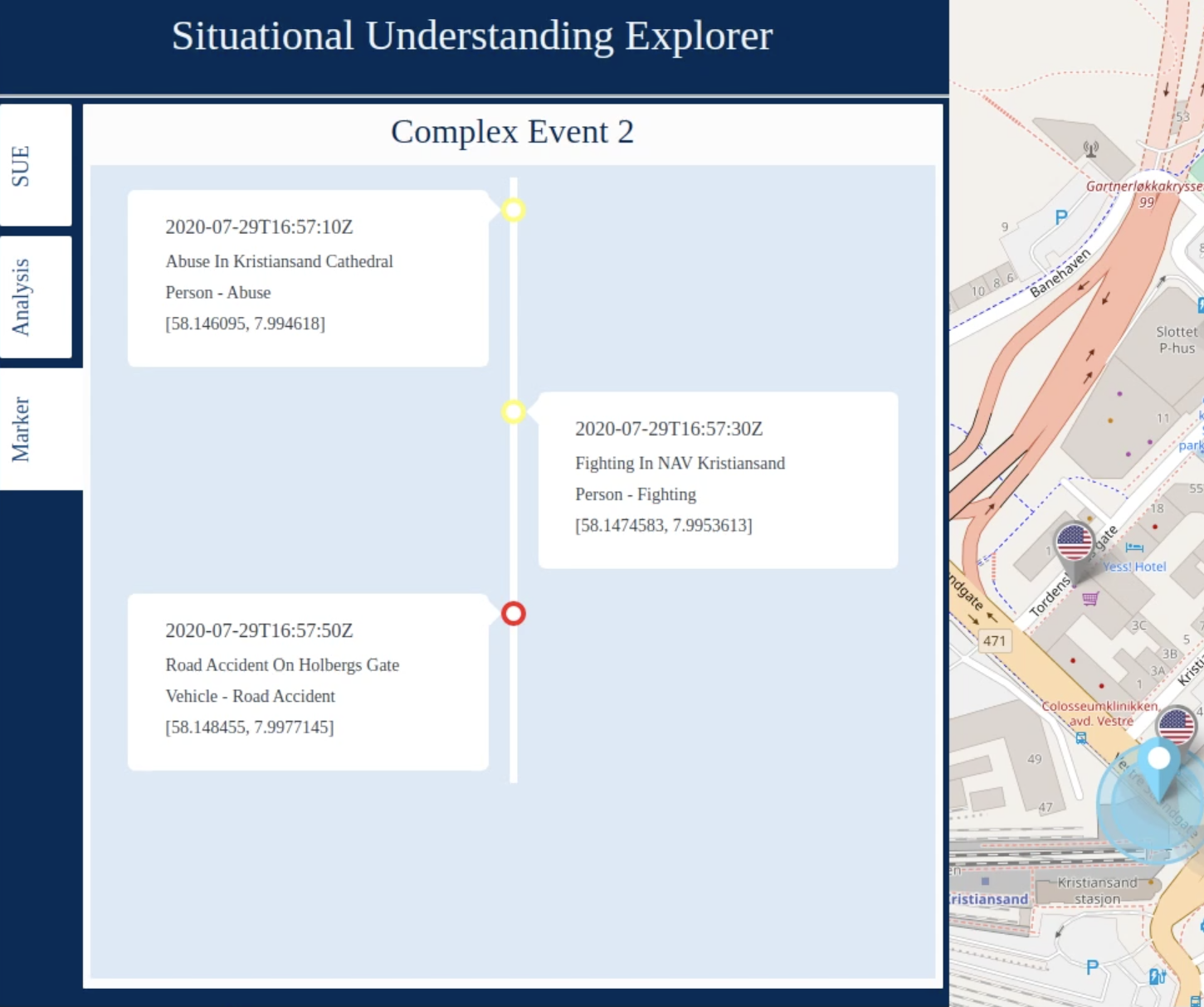}
\caption{Details of the complex event (previously seen on the map in Figure~\ref{fig:sue1}) visualised as a timeline.}
\label{fig:sue3}
\end{figure}

The main approach we use for complex event detection in our work is a neuro-symbolic combination of neural network services to detect simple events and a symbolic event calculus-based reasoner to identify interrelated complex events~\cite{Vilamala:2019}. The previously-seen example in Figure~\ref{fig:sue1} showed a detected complex event (the `!' marker) consisting of three simple events on the map. Details of this event are available on the marker tab as shown in Figure~\ref{fig:sue3}. Details of the event calculus-based reasoning are also available as a symbolic explanation.

\section{Current Status}

The SUE platform has been used as a front-end for two integrated demonstrations of CSU research capabilities comprising parts of the Distributed Analytics and Information Science International Technology Alliance (DAIS ITA)\footnote{http://sl.dais-ita.org/science-library/projects}. These demonstrations included the processing of real audiovisual sensor date previously collected as part of live exercises, streamed into SUE via its WebSocket interface. Two CSU services are currently integrated into SUE via WebSocket endpoints and JSON-based messages: (1) multimodal activity recognition incorporating the selective relevance technique~\cite{Taylor:2020}, which in turn builds on previous work including layerwise relevance propagation (LRP)~\cite{Bach:2015} for 3D convolutional neural networks~\cite{Tran:2015} trained on video activity recognition tasks \cite{Soomro:2012}; (2) complex event processing~\cite{Vilamala:2019} based on probabilistic logic programming via ProbLog~\cite{deraedt:ijcai07,fierens:tplp15} and the event calculus via ProbEC~\cite{Skarlatidis2015}. Integration of the latter with a third service based on the FastLAS scalable inductive logic programming system~\cite{Law:2020} for learning complex event detection rules has also been demonstrated via SUE.

\section{Conclusion and Future Work}

This paper has presented an experimentation platform for coalition situational understanding research that highlights capabilities in explainable artificial intelligence/machine learning (AI/ML) and integration of symbolic and subsymbolic AI/ML approaches for event processing. The platform is open and lightweight, to facilitate experiments and demonstrations. Its features are tailored to coalition multi-domain operations with emphasis on asset interoperability and ad hoc human-machine teaming in a dense urban terrain setting. The platform is easily extensible to incorporate non-spatial data sources such as social media, and on-the-spot reports from human operatives (e.g., patrols or guards). Current work is geared towards technology integration experiments; planned future work will involve human participant experiments to measure the utility of layered explanations in ad hoc coalition teaming tasks.


\end{document}